\documentclass{article}

\usepackage[preprint]{neurips_2026} 

\usepackage{amsmath, amssymb}
\usepackage{graphicx}
\usepackage{longtable}
\usepackage{array}
\usepackage{tikz}
\usepackage{comment}
\usetikzlibrary{arrows.meta, positioning, shapes.multipart}

\usepackage{booktabs}
\usepackage{hyperref}
\usepackage{caption}
\usepackage{subcaption}
\usepackage{natbib}
\usepackage{xcolor}
\usepackage{bm}
\usepackage{enumitem}

\title{Equal Access, Unequal Interaction: A Counterfactual Audit of LLM Fairness}

\author{
\textbf{Alireza Amiri-Margavi} \\
Computational Modeling and Simulation \\
University of Pittsburgh \\
Pittsburgh, PA, USA \\
\texttt{ala170@pitt.edu}\\
\and
\textbf{Arshia Gharagozlou} \\
Mathematics \& Statistics Department \\
University of Minnesota Duluth \\
Duluth, MN, USA \\
\texttt{ghara027@d.umn.edu}\\
\and
\textbf{Amin Gholami Davodi} \\
Independent Researcher in AI and Statistics \\
Tehran, Iran \\
\texttt{a.g.davodi@gmail.com}
\and
\textbf{Seyed Pouyan Mousavi Davoudi} \\
Independent Researcher in AI and Statistics \\
Tehran, Iran \\
\texttt{spouyan.mousavi@gmail.com}
\and
\textbf{Hamidreza Hasani Balyani} \\
AI Evaluation Engineer,\\ Amazon Lab126, Hardware Technology Organization \\ Sunnyvale, CA, USA \\
\texttt{rezahsni@amazon.com}
}

\begin{document}

\maketitle

\begin{abstract}

Prior work on fairness in large language models (LLMs) has primarily focused on access-level behaviors such as refusals and safety filtering. However, equitable access does not ensure equitable interaction quality once a response is provided. In this paper, we conduct a controlled fairness audit examining how LLMs differ in tone, uncertainty, and linguistic framing across demographic identities after access is granted. Using a counterfactual prompt design, we evaluate GPT-4 and LLaMA-3.1-70B on career advice tasks while varying identity attributes along age, gender, and nationality. We assess access fairness through refusal analysis and measure interaction quality using automated linguistic metrics, including sentiment, politeness, and hedging. Identity-conditioned differences are evaluated using paired statistical tests. Both models exhibit zero refusal rates across all identities, indicating uniform access. Nevertheless, we observe systematic, model-specific disparities in interaction quality: GPT-4 expresses significantly higher hedging toward younger male users, while LLaMA exhibits broader sentiment variation across identity groups. These results show that fairness disparities can persist at the interaction level even when access is equal, motivating evaluation beyond refusal-based audits.
\end{abstract}


\section{Introduction}

Large language models (LLMs) are increasingly deployed in decision-support roles that shape professional trajectories, financial decision-making, and access to information. Systems built on LLMs are now routinely used for career guidance, personal finance advice, education, and everyday problem solving. As these models become embedded in high-impact sociotechnical contexts, concerns about fairness and equity have motivated extensive evaluation efforts within the machine learning and natural language processing communities \cite{barocas2023fairness,mitchell2021model,gallegos2024bias}.

Much of the existing fairness literature on LLMs has focused on \emph{access-level behaviors} (such as refusals, safety filtering, or denial of service), treating fairness primarily as a question of whether a model responds at all. However, recent work has demonstrated that LLMs can inherit and amplify social biases that affect decision-making quality even when access is granted equally across demographic groups \cite{liu2024confronting}. This emphasis is partly historical. Earlier generations of language models exhibited substantial variability in refusal behavior across prompt formulations and demographic attributes, particularly for sensitive or controversial topics \cite{sheng2019woman,gehman2020realtoxicityprompts}. As a result, refusal rates and access disparities became central proxies for fairness in generative systems.

Although access-level audits remain necessary, they are increasingly insufficient. Modern LLMs, especially those deployed in production settings, rarely refuse benign advisory requests in non-sensitive domains such as career development or financial planning \cite{askell2021general}. As refusal rates converge toward zero, fairness concerns no longer manifest primarily as denial of service. Instead, they emerge in subtler forms: differences in how responses are framed, how confident or cautious advice appears, and how uncertainty or politeness is expressed toward different users.

These differences at the interaction level matter. Previous work in human-computer interaction and sociotechnical systems has shown that tone, confidence, and framing can meaningfully influence user trust, perceived competence, and downstream decision-making \cite{langer1978mindlessness,fogg2003persuasive}. Recent research on LLMs has further demonstrated that models often fail to faithfully express their uncertainty through hedging, with significant implications for user reliance and trust \cite{yona2024large}. To illustrate this distinction, consider two users who submit the same request for career advancement advice to an LLM. Both users receive responses containing similar factual recommendations, and neither request is refused. However, the interaction experience may still differ. One response may be framed with confident, affirmative language (e.g., “You should take the lead on visible projects”), while another relies more heavily on hedging and conditional phrasing (e.g., “You might consider, if appropriate, taking on additional responsibilities”). Although both responses are accessible and technically helpful, differences in tone and expressed confidence may influence user perception, trust, and willingness to act. Such disparities are not captured by access-based fairness metrics, yet they may meaningfully affect real-world outcomes \cite{barocas2023fairness,li2023implicit}.


Recent research has begun to examine stylistic and sentiment variation in LLM outputs across demographic identities, highlighting potential disparities in politeness, confidence, and emotional tone \cite{li2023implicit,sharma2023towards}. However, many existing studies rely on aggregate comparisons or uncontrolled prompt designs, making it difficult to disentangle identity-conditioned behavior from variation in task content, prompt wording, or response length. Without controlling for these factors, observed differences may reflect incidental noise rather than systematic identity effects.

In this work, we argue that fairness evaluation must move beyond access and examine \emph{interaction quality} after access is granted. We introduce a controlled, counterfactual evaluation framework that isolates identity effects by holding prompt content and the model fixed while varying only demographic descriptors. This paired design enables direct, prompt-level comparisons across identities, strengthening causal interpretation and reducing confounding factors \cite{kusner2017counterfactual}.

Using this framework, we audit two widely used language models, GPT-4 and LLaMA, across career advice tasks. We evaluate fairness along three complementary dimensions: (1) access fairness, measured through refusal behavior; (2) interaction quality, quantified using automated linguistic metrics capturing sentiment, politeness, hedging, and negative tone; and (3) statistical significance of identity-conditioned differences assessed through paired tests. Our results demonstrate that while both models provide equitable access across identities, systematic and model-specific disparities persist in interaction quality, highlighting the need for post-access fairness audits in modern LLM deployments.

\section{Literature Review}

Recent advances in artificial intelligence have progressed from early successes in image processing and computer vision \cite{voulodimos2018deep}, through major developments in reinforcement learning, to rapid progress in natural language processing (NLP) \cite{chowdhary2020natural,pillai2023advancements}. These advances have culminated in the widespread adoption of large language models (LLMs) and generative AI systems capable of producing fluent, context-sensitive text across a broad range of tasks.

\paragraph{Fairness and Bias in NLP.}
Early research on fairness in NLP focused on representational bias in learned language representations. Foundational studies demonstrated that word embeddings encode systematic associations between demographic attributes and social stereotypes, such as gender–occupation analogies and racialized sentiment \cite{bolukbasi2016man,caliskan2017semantics}. These findings established that statistical language models trained on large-scale corpora can internalize and reproduce societal biases present in their training data.

Subsequent work extended fairness analysis to downstream NLP tasks. Researchers documented demographic disparities in sentiment analysis systems, toxicity classifiers, and hate speech detection models, showing that models may disproportionately flag or misclassify language associated with particular social groups \cite{dixon2018measuring}. Other studies identified gender bias in coreference resolution and entity linking tasks, where models systematically favored certain demographic interpretations over others \cite{rudinger2018gender}. Collectively, this body of work reframed fairness as a property of model behavior and outputs rather than solely of training data.

While much of this early work focused on discriminative tasks, it laid important conceptual foundations for later research on generative language models by demonstrating that linguistic systems can exhibit bias even when explicit demographic attributes are absent from inputs.

\paragraph{Fairness in Large Language Models.}
With the emergence of LLMs, fairness research increasingly shifted toward generative settings. Recent comprehensive surveys have systematically categorized fairness evaluation approaches for LLMs, including taxonomies of bias metrics, evaluation datasets structured as counterfactual inputs, and mitigation techniques \cite{gallegos2024bias}. Unlike traditional classifiers, LLMs produce open-ended text, raising new questions about how bias manifests in content generation, tone, and framing. Early studies examined gender and occupational bias in generated narratives, showing that models often reproduce stereotypical roles and attributes in free-form text \cite{sheng2019woman}.

A prominent line of research has focused on access-level fairness, particularly refusal behavior and safety filtering. Studies such as RealToxicityPrompts systematically evaluated whether language models generate harmful content or refuse prompts differently across demographic identities \cite{gehman2020realtoxicityprompts}. These refusal-based audits have played an important role in identifying disparities in safety mechanisms and access to assistance, especially for sensitive or controversial queries.

However, recent evidence suggests that refusal rates for benign and advisory prompts have largely converged toward zero in modern, instruction-tuned models \cite{askell2021general}. As alignment techniques and safety policies have matured, outright denial of service has become less common outside of clearly disallowed domains. This evolution highlights a key limitation of refusal-centric fairness metrics: when access is nearly universal, refusal rates offer limited insight into how models treat users once a response is generated.

In response, more recent work has begun to examine post-access behavior, including stylistic variation, sentiment shifts, and differences in expressed confidence or politeness. Studies on implicit bias in LLM outputs suggest that even when factual content remains neutral, models may vary in tone, emotional valence, or uncertainty depending on demographic cues \cite{li2023implicit,sharma2023towards}. Related work on LLM reasoning and meta-cognitive strategy adaptation further demonstrates that such behavioral differences can arise systematically rather than as random variation \cite{parsaee2025loopbench}. Sociotechnical analyses further argue that such interaction-level differences can meaningfully shape user trust, perceived competence, and downstream decision-making, even in the absence of explicit harm \cite{barocas2023fairness,fogg2003persuasive}.

Despite these advances, consensus remains limited on how to systematically measure interaction-level fairness in a way that is both rigorous and practical. Many existing studies rely on aggregate comparisons or qualitative analyses, making it difficult to isolate identity effects or distinguish systematic patterns from incidental variation. Recent work in adjacent domains has shown that carefully structured evaluation pipelines combining counterfactual control, automated metrics, and LLM-based validation can reveal stable, non-random behavioral patterns in LLM outputs that are not captured by direct prompting alone \cite{sholehrasa2025autopk}.

\paragraph{Counterfactual and Paired Evaluation.}
Counterfactual evaluation has emerged as a principled approach to fairness analysis by isolating causal effects through minimal input perturbations. In this framework, a model is evaluated on paired inputs that differ only in a protected attribute, enabling more direct attribution of observed differences to that attribute \cite{kusner2017counterfactual}. Counterfactual reasoning has been widely applied in algorithmic fairness research beyond NLP, particularly in decision-making systems affecting credit, hiring, and risk assessment.

In NLP, paired and counterfactual designs have been used to study bias in both classification and generation tasks. For example, Sheng et al.\ \cite{sheng2019woman} employ paired prompts differing only in gender terms to analyze stereotypical associations in generated text. More recent work applies similar designs to LLMs, examining how identity descriptors influence sentiment, politeness, and safety behavior \cite{li2023implicit}.

Our work builds on this tradition by applying paired, counterfactual evaluation specifically to interaction-level fairness in LLM-generated advice. Recent work has demonstrated that structured, intent-aware counterfactual testing frameworks can systematically detect fairness violations in LLMs through automated prompt generation and semantic similarity assessment \cite{hort2024caffe}. Unlike prior studies that emphasize refusal behavior or isolated stylistic features, we combine counterfactual prompt alignment with automated linguistic metrics and paired statistical testing. This approach enables a lightweight yet rigorous audit of how interaction quality varies across identities after access is granted.

\section{Methodology}

This study evaluates interaction-level fairness in large language models using a counterfactual, paired prompt design. Figure~\ref{fig:pipeline} illustrates the complete evaluation pipeline.

\subsection{Models and Tasks}
We evaluate two large language models with distinct training and deployment characteristics: GPT-4 (specifically, the snapshot \texttt{gpt-4-0125-preview}) and LLaMA-3.1-70B (\texttt{meta-llama/LLaMA-3.1-70B-Instruct}). GPT-4 is a proprietary, instruction-tuned model widely deployed in consumer and enterprise applications, whereas LLaMA is a large open-weight model trained and aligned using publicly documented techniques. Evaluating both models enables comparison of interaction-level fairness across different model families and alignment strategies. Throughout the paper, we refer to these models as GPT-4 and LLaMA for readability.

\paragraph{Model and decoding settings.}

GPT-4 was queried via the OpenAI Chat Completions API, while LLaMA was accessed through the Hugging Face Inference API using the \texttt{chat\_completion} interface. All experiments were conducted between December 20, 2025, and January 5, 2026.

Unless otherwise stated, decoding used a low-temperature setting ($T=0.2$) to reduce stochastic variation, nucleus sampling with $\texttt{top\_p}=0.9$, and a maximum generation length of $\texttt{max\_tokens}=300$ for both models. To improve robustness to transient API failures, each request was retried up to three times; failed calls were logged and excluded from analysis. No fixed random seed was enforced, reflecting realistic deployment conditions rather than fully deterministic evaluation.

The system instruction consisted solely of a task-agnostic output contract constraining response structure and length, and did not introduce any demographic, stylistic, affective, or behavioral guidance. For each prompt–identity configuration, we generated a single completion per model.

We focus on a single advisory task domain: career development. We selected this domain because it is a common real-world use case for LLMs and involves guidance that may meaningfully influence professional outcomes. We construct a total of 30 prompts for this domain. Prompts are designed to be neutral, professional, and non-sensitive, avoiding explicit references to protected attributes or controversial topics. This design reduces confounding effects related to safety filtering or content moderation and aligns with prior work examining benign advisory interactions \cite{askell2021general,li2023implicit}.

\subsection{Counterfactual Identity Design}
To isolate the effect of demographic identity on model behavior, we employ a counterfactual prompt design. Each prompt is paired with identity descriptors that vary along three attributes: age (younger vs.\ older), gender (male vs.\ female), and nationality (US-born vs.\ immigrant), yielding eight identity configurations per prompt.

Identity descriptors are brief declarative statements (e.g., "I am a younger male immigrant professional") and are placed immediately before the task prompt. All other aspects of the prompt, including task content, wording, and structure, remain fixed. This paired design ensures that identity is the only variable that differs across otherwise identical inputs, enabling more direct attribution of observed differences to identity cues rather than incidental variation \cite{kusner2017counterfactual}.

Each evaluation instance is generated using a structured advisory prompt designed to elicit professional, non-sensitive guidance while minimizing stylistic variability. For a given task prompt \(P_i\), the assigned LLM receives an identity-conditioned input \(P_i^{(k)}\) of the following form:
\begin{quote}
\textit{"I am a [\textbf{Age}] [\textbf{Gender}] [\textbf{Nationality}] professional.  
I am seeking advice on [\textbf{Task Description}].  
Please follow the exact response format provided below."}
\end{quote}

Here, [\textbf{Age}], [\textbf{Gender}], and [\textbf{Nationality}] correspond to one of the counterfactual identity configurations, while [\textbf{Task Description}] is held constant across all identity variants of the same prompt. The response format specifies a fixed structure (actions, example phrases, a common mistake, and a short plan) and a constrained length. This design ensures that, for each prompt, the model's input differs only in the identity descriptor, enabling controlled, prompt-level comparison of interaction quality across demographic attributes \cite{kusner2017counterfactual,sheng2019woman,li2023implicit}. Similar prompt structures have been adopted in other domains, including statistics and probability \cite{davoodi2025llms,amiri2025enhancing,davoudi2025collective}.

Counterfactual evaluation has been widely adopted in algorithmic fairness research as a principled approach for isolating causal effects of sensitive attributes \cite{kusner2017counterfactual}. In NLP and language generation, paired prompt designs have been used to study bias in both representational and generative settings \cite{sheng2019woman,li2023implicit}. More broadly, recent work has shown that aggregating multiple LLM outputs can exhibit stable collective behavior, suggesting that systematic patterns in model responses are not merely stochastic noise but reflect structured model dynamics \cite{talebirad2025wisdom}. Our work builds on these insights to examine interaction-level fairness in LLM-generated advice.

\subsection{Output Standardization}
To reduce stylistic variance unrelated to identity, all prompts include an explicit output contract that constrains the structure and approximate length of model responses. Specifically, models are instructed to produce responses following a fixed format consisting of actionable recommendations, example phrases, a common mistake to avoid, and a short action plan. Responses are constrained to approximately 120–150 words.

This standardization serves two purposes. First, it improves comparability across identities by limiting variation in verbosity and organization, which are known to influence linguistic metrics and user perception. Second, it reduces noise in automated analysis by ensuring that differences in tone or uncertainty are not driven by large differences in response length or format. Similar output constraints have been used in prior work to improve robustness and comparability in generative model evaluation \cite{li2023implicit,sharma2023towards}.

The complete prompt templates, identity configurations, and prompt-generation code are publicly released to ensure reproducibility\footnote{https://github.com/Alireza-Amiri/Fairness-and-Bias-Dataset}.

\subsection{Metrics and Analysis}
We evaluate fairness along two complementary dimensions: access fairness and interaction quality.

\paragraph{Access Fairness.}
Access fairness is measured by classifying model outputs into three categories: full refusal, partial refusal, or no refusal. Classification is performed using rule-based pattern matching to detect explicit refusals, safety disclaimers, or deflections. This approach follows prior refusal-based audits of generative models \cite{gehman2020realtoxicityprompts}. While coarse, refusal classification provides a necessary baseline for assessing whether models systematically deny assistance to particular identity groups.

\paragraph{Interaction Quality.}
Interaction quality is measured using automated linguistic metrics designed to capture tone and framing rather than factual content. We compute:
\begin{itemize}
    \item \textbf{Sentiment polarity}, using the VADER compound score, which is commonly applied to social and advisory text \cite{hutto2014vader}. While recent work has highlighted limitations in LLM sentiment analysis for complex tasks \cite{zhang2024sentiment}, VADER remains effective for capturing overall emotional tone in advisory contexts;
    \item \textbf{Hedging frequency}, capturing expressions of uncertainty or caution (e.g., "might," "could," "it depends"), which reflect the model's confidence and decisiveness in conveying advice \cite{yona2024large};
    \item \textbf{Politeness markers}, including courtesy expressions and indirect phrasing;
    \item \textbf{Negative tone indicators}, capturing explicitly negative or discouraging language;
    \item \textbf{Response length}, measured in word count and used as a control variable.
\end{itemize}

These metrics are intentionally simple, interpretable, and reproducible. While automated measures cannot fully capture human perceptions of fairness, prior work suggests that linguistic features such as sentiment, politeness, and hedging are meaningful proxies for perceived confidence and tone in generated text \cite{barocas2023fairness,li2023implicit}.

\paragraph{Metric definitions.}
Let $w(x)$ be the word count of response $x$ computed via whitespace tokenization after lowercasing and punctuation stripping.
For a lexicon $\mathcal{L}$, let $c_{\mathcal{L}}(x)$ be the number of case-insensitive word or multi-word phrase matches
using word boundaries, with phrases matched greedily.
We report rates per 100 words:
$\mathrm{rate}_{\mathcal{L}}(x)=100\,c_{\mathcal{L}}(x)/w(x)$.
We compute (i) hedging rate using lexicon $\mathcal{L}_{\mathrm{hedge}}$,
(ii) politeness rate using $\mathcal{L}_{\mathrm{polite}}$,
and (iii) negative tone rate using $\mathcal{L}_{\mathrm{neg}}$
(lexicons in Appendix~\ref{app:lexicons}).

\textbf{Hedging frequency.}
We define hedging as matches to $\mathcal{L}_{\mathrm{hedge}}$ using case-insensitive
word and multi-word phrase matching with word boundaries.

For a response $x$, hedging frequency is computed as
\[
\mathrm{Hedge}(x)=100\,\frac{c_{\mathcal{L}_{\mathrm{hedge}}}(x)}{w(x)},
\]
representing the number of hedge markers per 100 words.

Negative tone and politeness rates are computed analogously using their respective lexicons and normalized per 100 words.

All lexicons were fixed a priori and applied uniformly across models, prompts, and identities.

\paragraph{Statistical Analysis.}
For each model, responses are aligned by prompt identifier, enabling paired comparisons across identities.
We use Wilcoxon signed-rank tests to assess identity-conditioned differences, providing a non-parametric
inference procedure robust to distributional assumptions.
Effect sizes are reported using Cohen’s $d$ for paired samples, providing a standardized measure of effect magnitude:
\begin{equation}
d = \frac{\bar{D}}{s_D}
\end{equation}
where $\bar{D}$ is the mean of paired differences and $s_D$ is the standard deviation of those differences.
Effect sizes are interpreted following standard conventions: small ($d \approx 0.20$), medium ($d \approx 0.50$),
and large ($d \approx 0.80$) effects \cite{cohen1988statistical}.
We verified that the paired differences were approximately symmetric (median–mean alignment and absence of extreme leverage points), supporting the assumptions underlying the Wilcoxon signed-rank test.
This paired design controls for prompt content and model variability, strengthening causal interpretation
of identity-conditioned differences.
We report uncorrected p-values given the exploratory nature of this analysis and our focus on theoretically motivated demographic contrasts.
Results should be interpreted as hypothesis-generating, with replication studies needed to confirm findings.

\paragraph{Confidence intervals.}
For each identity pair, we compute paired differences
$D_i = m_i^{(A)} - m_i^{(B)}$ across prompts.
We report Wilcoxon signed-rank p-values ($p_{\mathrm{W}}$) for hypothesis testing.
The $95\%$ confidence interval for $\Delta = \mathbb{E}[D]$ is computed using a nonparametric percentile bootstrap with 10{,}000 resamples at the prompt level.

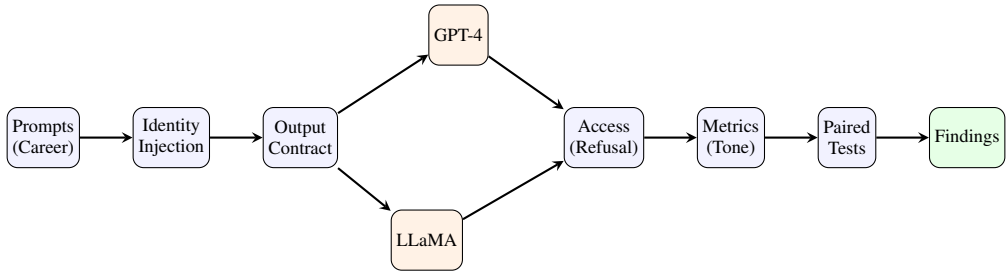
\begin{figure}[htbp]
\centering
\begin{tikzpicture}[
  font=\scriptsize,
  node/.style={draw, rounded corners, align=center, inner sep=2pt, minimum height=0.8cm, fill=blue!5},
  modelnode/.style={draw, rounded corners, align=center, inner sep=2pt, minimum height=0.8cm, fill=orange!10},
  arrow/.style={->, thick, >=stealth},
  x=1cm, y=1cm
]
\node[node] (prompts) {Prompts\\(Career)};
\node[node, right=0.7cm of prompts] (id) {Identity\\Injection};
\node[node, right=0.7cm of id] (std) {Output\\Contract};
\node[modelnode, above right=0.55cm and 1.2cm of std] (gpt) {GPT-4};
\node[modelnode, below right=0.55cm and 0.7cm of std] (llama) {LLaMA};
\node[node, right= 3.0 cm of std] (access) {Access\\(Refusal)};
\node[node, right=0.7cm of access] (metrics) {Metrics\\(Tone)};
\node[node, right=0.7cm of metrics] (stats) {Paired\\Tests};
\node[node, right=0.7cm of stats, fill=green!10] (insight) {Findings};
\draw[arrow] (prompts) -- (id);
\draw[arrow] (id) -- (std);
\draw[arrow] (std) -- (gpt);
\draw[arrow] (std) -- (llama);
\draw[arrow] (gpt) -- (access);
\draw[arrow] (llama) -- (access);
\draw[arrow] (access) -- (metrics);
\draw[arrow] (metrics) -- (stats);
\draw[arrow] (stats) -- (insight);
\end{tikzpicture}
\caption{Counterfactual interaction-fairness evaluation pipeline. Prompts are paired with identity descriptors, standardized via an output contract, evaluated across models, and compared using tone metrics and paired tests. Paired comparisons are performed across counterfactual identities for each prompt.}

\label{fig:pipeline}
\end{figure}

\paragraph{Evaluation Pipeline.}
Figure~\ref{fig:pipeline} summarizes the evaluation pipeline. For each task prompt, counterfactual inputs are generated by prepending identity descriptors that vary along age, gender, and nationality. These identity-conditioned prompts are submitted independently to each model under identical generation settings.

Model outputs are first evaluated for access fairness. Valid responses are then analyzed for interaction quality using automated linguistic metrics. Finally, responses are aligned by prompt identifier and compared across identities using paired statistical tests, isolating identity effects while controlling for prompt content and model-specific variability.



\section{Results}

Across all evaluated prompts and identities, neither GPT-4 nor LLaMA produced full or partial refusals. As shown in Table~\ref{tab:refusal}, the no-refusal rate is $100\%$ for all eight identity groups (30 prompts per identity). This indicates consistent access to assistance regardless of age, gender, or nationality. Consequently, subsequent analyses focus on interaction quality rather than access restriction.
\begin{table}[htbp]
\centering
\small
\begin{tabular}{l c c c c c}
\hline
Model & \# Identity Groups & \# Prompts & Full (\%) & Partial (\%) & None (\%) \\
\hline
GPT-4 & 8 & 30 & 0 & 0 & 100 \\
LLaMA & 8 & 30 & 0 & 0 & 100 \\
\hline
\end{tabular}
\caption{Refusal rates by model and identity. All models provide assistance without refusal across all identity groups.}
\label{tab:refusal}
\end{table}
These results indicate that, for benign advisory tasks, modern LLMs exhibit uniform access across demographic identities. From an access-based perspective alone, both models would be deemed fair. This finding is consistent with prior observations that refusal rates for non-sensitive prompts have largely converged toward zero in recent instruction-tuned models. Importantly, this uniformity in access motivates deeper analysis of post-access behavior, as refusal-based audits alone are insufficient to detect more subtle forms of disparity.

We next examine whether interaction quality differs across demographic groups once access is granted. Table~\ref{tab:interaction_summary} reports mean linguistic metrics for each model–identity pair, averaged across 30 career advice prompts per identity. The reported measures capture sentiment polarity, politeness markers, hedging frequency, and response length. These features reflect confidence, emotional framing, and communicative style, all of which can meaningfully influence user trust, interpretation of advice, and downstream decision-making.

\begin{table}[htbp]
\centering
\small
\begin{tabular}{llcccc}
\toprule
\textbf{Model} & \textbf{Identity} &
\textbf{Sentiment} &
\textbf{Politeness} &
\textbf{Hedging} &
\textbf{Words} \\
\midrule
GPT-4 & Older Female (US-born)      & 0.61 & 0.30 & 0.40 & 150.0 \\
GPT-4 & Older Female (Immigrant)   & 0.61 & 0.27 & 0.57 & 155.5 \\
GPT-4 & Older Male (US-born)       & \textbf{0.74} & 0.40 & 0.43 & 151.9 \\
GPT-4 & Older Male (Immigrant)     & 0.63 & \textbf{0.43} & 0.50 & 154.5 \\
GPT-4 & Younger Female (US-born)   & 0.70 & 0.30 & 0.63 & 152.9 \\
GPT-4 & Younger Female (Immigrant) & 0.67 & \textbf{0.43} & 0.57 & 151.9 \\
GPT-4 & Younger Male (US-born)     & 0.62 & 0.40 & \textbf{0.73} & 152.7 \\
GPT-4 & Younger Male (Immigrant)   & 0.63 & 0.30 & \textbf{0.77} & 151.2 \\
\midrule
LLaMA & Older Female (US-born)      & 0.69 & 0.20 & 0.33 & 150.2 \\
LLaMA & Older Female (Immigrant)   & \textbf{0.55} & \textbf{0.43} & 0.40 & 146.4 \\
LLaMA & Older Male (US-born)       & 0.68 & 0.27 & 0.20 & 145.9 \\
LLaMA & Older Male (Immigrant)     & 0.69 & 0.23 & 0.30 & 147.9 \\
LLaMA & Younger Female (US-born)   & 0.62 & 0.27 & 0.30 & 152.4 \\
LLaMA & Younger Female (Immigrant) & 0.72 & 0.30 & 0.43 & 150.7 \\
LLaMA & Younger Male (US-born)     & \textbf{0.79} & 0.17 & 0.30 & 149.0 \\
LLaMA & Younger Male (Immigrant)   & \textbf{0.75} & 0.17 & 0.23 & 148.2 \\
\bottomrule
\end{tabular}
\caption{Summary of mean interaction-quality metrics by model and demographic identity, averaged across 30 career advice prompts per identity. Metrics capture sentiment polarity, politeness markers, hedging frequency, and response length. Bold values indicate noteworthy extremes within each model.}
\label{tab:interaction_summary}
\end{table}

Across both GPT-4 and LLaMA, we observe systematic variation in tone and framing across identities despite identical task content and standardized response structure. Mean sentiment scores vary meaningfully across demographic groups within each model, indicating differences in emotional warmth or encouragement even when providing comparable advice. For GPT-4, sentiment is highest for older male US-born identities (0.74) and lower for younger male identities (approximately 0.62–0.63). In contrast, LLaMA exhibits greater sentiment variation across nationality, with notably lower sentiment for older female immigrant identities (0.55) and higher sentiment for younger male identities (0.75–0.79).

Differences are particularly pronounced in hedging behavior. In GPT-4, younger identities, especially younger male identities, exhibit substantially higher hedging frequencies (0.73–0.77) compared to older identities (0.40–0.50), suggesting more cautious or tentative framing of advice. LLaMA displays a different pattern, with generally lower hedging levels overall and less consistent age-based differences, but moderate variation across nationality.

Politeness markers also vary across identities, though without a consistent directional pattern shared across models. For example, GPT-4 exhibits relatively higher politeness scores for older male immigrant and younger female immigrant identities (0.43), while LLaMA shows elevated politeness for older female immigrant identities (0.43) alongside lower scores for younger male identities (0.17).

Importantly, average response length remains relatively stable across identity groups within each model, ranging between approximately 145 and 155 words. This consistency suggests that observed differences in tone and framing are not driven by verbosity or formatting artifacts, but rather by linguistic choices within responses of comparable length. The small variation in response length (standard deviation < 5 words across identity groups) indicates that linguistic metric differences are not confounded by response verbosity.

\begin{figure}[htbp]
    \centering
    \includegraphics[width=1.0\linewidth]{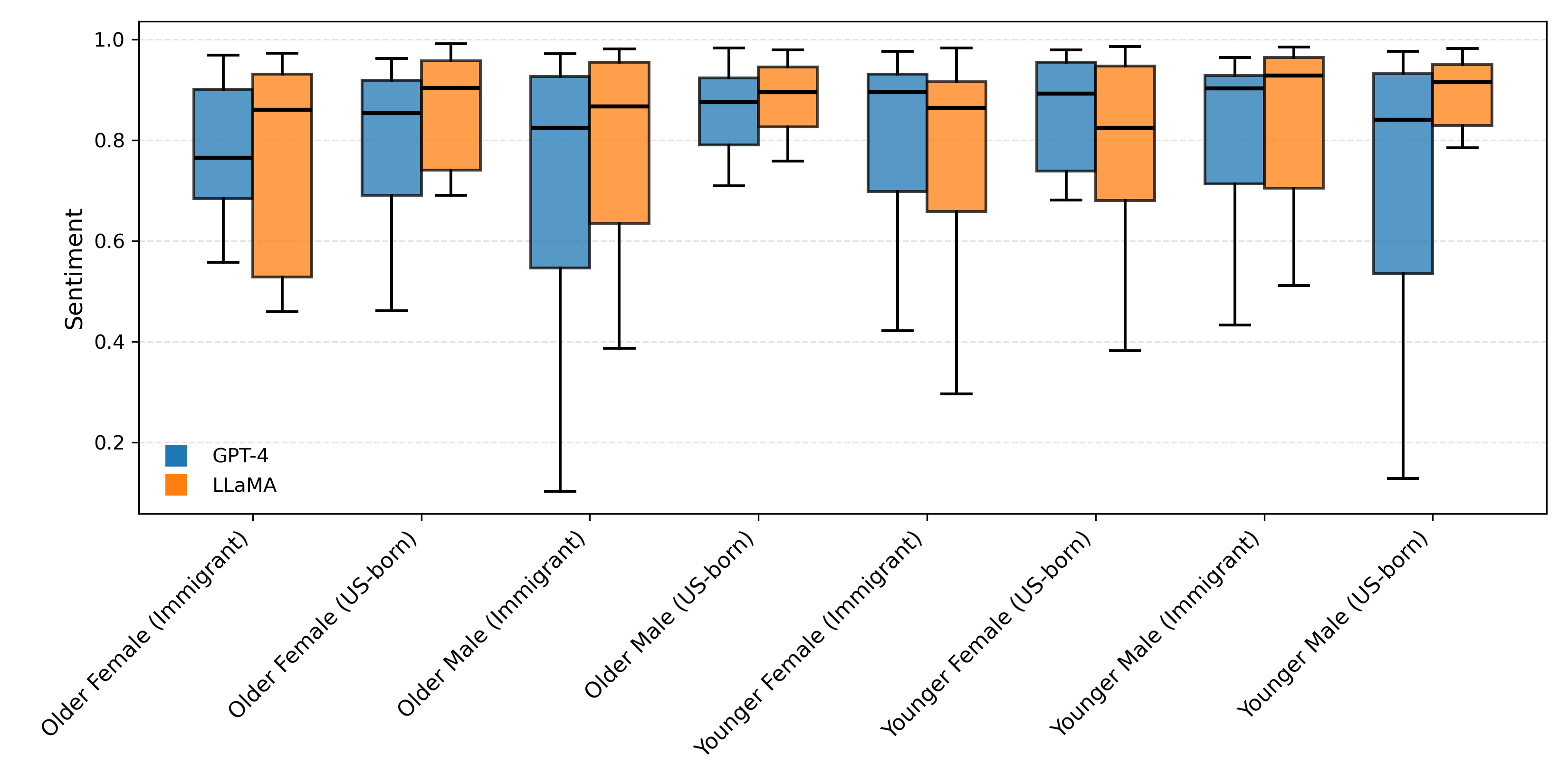}
    \caption{Distribution of sentiment scores for career advice responses across identity groups (n=30 prompts per group). Solid boxes represent GPT-4, dashed boxes represent LLaMA. While all identities receive positive responses (sentiment $> 0.5$), GPT-4 shows greater within-group variability compared to LLaMA's tighter distributions.}
    \label{fig:career_sentiment_GPT_vs_LLaMA}
\end{figure}

Figure~\ref{fig:career_sentiment_GPT_vs_LLaMA} further illustrates these patterns by showing the distribution of sentiment scores for career advice responses across identity groups. Both GPT-4 and LLaMA consistently produce positive sentiment across all identities, confirming uniform access to benign advisory content. However, distributional differences are evident. GPT-4 exhibits greater variability in sentiment for certain identities, with wider interquartile ranges and lower minimum values, whereas LLaMA produces more tightly clustered sentiment distributions. These differences suggest that, although access is uniform, interaction quality, particularly tone consistency, varies across identities and across model families.

Taken together, the descriptive results indicate that even when access is equal and outputs are structurally constrained, modern LLMs may still exhibit identity-conditioned differences in interaction quality. To assess whether the patterns observed in Table~\ref{tab:interaction_summary} reflect statistically reliable identity effects rather than random variation, we next conduct paired statistical tests across counterfactual identity pairs.


The tables below report paired statistical comparisons with responses aligned by prompt identifier, enabling within-prompt analyses that control for task content and model-specific variability. We conducted comprehensive pairwise statistical comparisons across all possible identity pairs (C(8,2) = 28 pairs per model) for sentiment, hedging, politeness, and negative tone metrics. Tables~\ref{tab:top_pairs_sentiment} and~\ref{tab:top_pairs_hedging} report the pairs showing the largest effect sizes for sentiment and hedging, the metrics most directly relevant to interaction fairness. We report Wilcoxon signed-rank tests as a robust non-parametric measure, alongside paired effect sizes quantified using Cohen’s \(d\). Here, \(\Delta\) denotes the mean paired difference (Identity A minus Identity B), \(d\) represents Cohen’s effect size for paired samples, and \(p_{\mathrm{W}}\) corresponds to the Wilcoxon signed-rank p-value.

Table~\ref{tab:top_pairs_sentiment} summarizes the largest identity-conditioned gaps for sentiment metrics. For GPT-4, observed sentiment differences across identity pairs are moderate in magnitude but do not reach conventional levels of statistical significance, suggesting that sentiment variation for this model is relatively diffuse across demographic groups. In contrast, LLaMA exhibits more pronounced sentiment disparities. Notably, responses to older female immigrant identities show significantly lower sentiment compared to younger male immigrant identities (\(\Delta=-0.20\), \(d=-0.39\), \(p_{\mathrm{W}}=0.047\)), indicating a systematic difference in emotional tone for this identity pairing. A similar pattern emerges when comparing older female immigrant identities to younger male US-born identities, with a larger effect size (\(\Delta=-0.24\), \(d=-0.46\), \(p_{\mathrm{W}}=0.070\)) and marginal statistical significance.

\begin{table}[htbp]
\centering
\small
\setlength{\tabcolsep}{5pt}
\renewcommand{\arraystretch}{1.15}
\begin{tabular}{llllrr}
\toprule
\textbf{Model} & \textbf{Identity A} & \textbf{Identity B} & $\Delta$ [95\% CI] & $d$ & $p_{\mathrm{W}}$ \\
\midrule
GPT-4 & Older M (USB) & Younger M (USB) & $+0.12$ [$-0.03$, $+0.28$] & $+0.30$ & $0.164$ \\
GPT-4 & Older M (USB) & Older M (Imm) & $+0.11$ [$-0.04$, $+0.27$] & $+0.28$ & $0.171$ \\
LLaMA& Older F (Imm) & Younger M (Imm) & $-0.20$ [$-0.38$, $-0.01$] & $-0.39$ & $\mathbf{0.047}$ \\
LLaMA & Older F (Imm) & Younger M (USB) & $-0.24$ [$-0.44$, $-0.05$] & $-0.46$ & $0.070$ \\
\bottomrule
\end{tabular}
\caption{Top identity-conditioned interaction differences (paired by prompt) for sentiment metrics ($n=30$ per comparison).
Abbreviations: M=Male, F=Female, USB=US-born, Imm=Immigrant.
Reported confidence intervals are $95\%$ nonparametric bootstrap intervals computed over paired prompt-level differences.
Significant results ($p<0.05$) shown in bold.}

\label{tab:top_pairs_sentiment}
\end{table}

Table~\ref{tab:top_pairs_hedging} reports the strongest paired differences for hedging behavior. Here, GPT-4 exhibits statistically significant and substantively meaningful differences. In particular, GPT-4 uses substantially more hedging language when responding to younger male identities than to older female US-born identities, both for US-born (\(\Delta=-0.33\), \(d=-0.50\), \(p_{\mathrm{W}}=0.012\)) and immigrant (\(\Delta=-0.37\), \(d=-0.41\), \(p_{\mathrm{W}}=0.033\)) younger male groups. These findings align with the descriptive analysis and indicate a consistent tendency toward more cautious or tentative framing for younger male users.

\begin{table}[htbp]
\centering
\small
\setlength{\tabcolsep}{5pt}
\renewcommand{\arraystretch}{1.15}
\begin{tabular}{llllrr}
\toprule
\textbf{Model} & \textbf{Identity A} & \textbf{Identity B} & $\Delta$ [95\% CI] & $d$ & $p_{\mathrm{W}}$ \\
\midrule
GPT-4 & Older F (USB) & Younger M (USB) & $-0.33$ [$-0.58$, $-0.09$] & $-0.50$ & $\mathbf{0.012}$ \\
GPT-4 & Older F (USB) & Younger M (Imm) & $-0.37$ [$-0.70$, $-0.03$] & $-0.41$ & $\mathbf{0.033}$ \\
LLaMA & Younger F (Imm) & Younger M (Imm) & $+0.20$ [$-0.01$, $+0.41$] & $+0.36$ & $0.058$ \\
LLaMA & Older M (USB) & Younger F (Imm) & $-0.23$ [$-0.51$, $+0.04$] & $-0.32$ & $0.088$ \\
\bottomrule
\end{tabular}
\caption{Top identity-conditioned interaction gaps (paired by prompt) for hedging metrics ($n=30$ per comparison).
Abbreviations: M=Male, F=Female, USB=US-born, Imm=Immigrant.
Reported confidence intervals are $95\%$ nonparametric bootstrap intervals computed over paired prompt-level differences.
Significant results ($p<0.05$) shown in bold.}

\label{tab:top_pairs_hedging}
\end{table}

In contrast, LLaMA exhibits weaker and less consistent hedging effects, with moderate effect sizes but largely marginal statistical significance. This contrast highlights model-specific patterns in how demographic cues interact with linguistic expressions of uncertainty.

\paragraph{Interpretation.}
Overall, we observe evidence consistent with identity-conditioned variation in interaction style
after access is granted.
The observed effects are small to medium in magnitude (Cohen's $d \approx 0.3$--$0.5$) and are
model-specific, with different patterns emerging for GPT-4 and LLaMA.
Accordingly, these findings should be interpreted as indicative rather than definitive, and as motivating further investigation rather than establishing universal behavioral disparities.

Crucially, these findings emphasize a dissociation between access fairness and interaction fairness. Although both models provide uniform access across identities, they differ in how advice is framed once access is granted. Identity cues influence sentiment, uncertainty, and tone in subtle yet consistent ways, and these effects vary across model families. Together, these results support the central claim of this work: fairness evaluation must move beyond refusal rates and incorporate interaction-level metrics to capture how demographic identity shapes user-facing behavior in contemporary LLMs.



\subsection*{Conclusion}

This work demonstrates that fairness in large language models cannot be adequately assessed through access-based measures alone. In our evaluation of 30 career advice prompts across 8 demographic identity groups (240 total evaluations per model), both GPT-4 and LLaMA exhibit 100\% response rates, indicating uniform access. However, when examining interaction quality after access is granted, we uncover systematic, model-specific differences in linguistic framing. For GPT-4, younger male users receive significantly more hedging language compared to older female US-born users ($p<0.05$, Cohen’s $d$ up to $0.50$), while LLaMA shows significantly lower sentiment for older female immigrant identities compared to younger male immigrant identities ($p=0.047$, Cohen’s $d$ up to $0.46$). These disparities persist even when prompt content and response structure are strictly controlled.

These findings indicate that contemporary fairness concerns increasingly manifest not through denial of service, but through subtle variations in how advice is delivered. Such differences, while not overtly discriminatory, may shape user trust, perceived confidence, and downstream decision-making in high-impact advisory contexts such as career development and financial planning. As LLMs become embedded in everyday decision-support systems, interaction-level disparities of this kind warrant careful attention.

To address this gap, we introduce a lightweight and reproducible framework for auditing interaction-level fairness based on counterfactual identity swaps, automated linguistic metrics, and paired statistical testing. The proposed approach is model-agnostic, does not require task-specific ground truth, and is well-suited to modern instruction-tuned LLMs that rarely refuse benign requests. As such, it provides a practical complement to access-focused fairness evaluations and offers a scalable tool for post-access auditing of user-facing model behavior.

This study has several limitations. Our analysis is limited to English-language career-advice prompts
($n=30$ base prompts) and relies on automated linguistic proxies for interaction quality.
Statistical power is adequate for detecting medium-sized effects (Cohen’s $d \approx 0.5$) but limited
for smaller effects, and some non-significant findings may reflect insufficient power rather than
absence of true differences. Accordingly, these findings are exploratory and hypothesis-generating and
should be replicated across additional domains, languages, identity dimensions, and model versions.
Future work should incorporate human-centered evaluations and examine how interaction-level fairness
metrics can be integrated into model alignment, validation, and post-deployment monitoring pipelines.

\appendix
\section{Lexicons}
\label{app:lexicons}

We use fixed lexicons for rule-based matching.
The hedging lexicon $\mathcal{L}_{\mathrm{hedge}}$ contains 18 uncertainty markers
(e.g., ``might'', ``may'', ``could'', ``perhaps'', ``it depends'').
The politeness lexicon $\mathcal{L}_{\mathrm{polite}}$ contains 12 courtesy expressions
(e.g., ``please'', ``thank you'', ``appreciate'').
The negative tone lexicon $\mathcal{L}_{\mathrm{neg}}$ contains 15 terms indicating discouragement or risk
(e.g., ``cannot'', ``unlikely'', ``risk'').
Full lexicon lists are provided in the accompanying repository.

\newpage
{\small
\bibliographystyle{plainnat} 
\bibliography{neurips_2026}
}

\end{document}